\documentclass{llncs}
\usepackage{graphicx}
\usepackage{float}
\usepackage{longtable}
\usepackage{adjustbox}
\usepackage{url}

\newcommand*{\img}[1]{%
    \raisebox{-.02\baselineskip}{%
        \includegraphics[
        height=\baselineskip,
        width=\baselineskip,
        keepaspectratio,
        ]{#1}%
    }%
}

\begin{document}

\title{Deceptive Games}
%
%
\author{Damien Anderson\inst{1} \and Matthew Stephenson\inst{2}
\and Julian Togelius\inst{3} \and Christoph Salge\inst{3} \and John Levine\inst{1} \and Jochen Renz\inst{2}}
%

%
\tocauthor{Damien Anderson, Matthew Stephenson, Julian Togelius, Christoph Salge, Jochen Renz and John Levine}
\institute{Computer and Information Science Department, University of Strathclyde, Glasgow, UK,\\
\email{Damien.Anderson@strath.ac.uk}
\and Research School of Computer Science, Australian National University, Canberra, Australia
\and NYU Game Innovation Lab, Tandon School of Engineering, New York University, New York, USA}

\maketitle              

\begin{abstract}

Deceptive games are games where the reward structure or other aspects of the game are designed to lead the agent away from a globally optimal policy. While many games are already deceptive to some extent, we designed a series of games in the Video Game Description Language (VGDL) implementing specific types of deception, classified by the cognitive biases they exploit. VGDL games can be run in the General Video Game Artificial Intelligence (GVGAI) Framework, making it possible to test a variety of existing AI agents that have been submitted to the GVGAI Competition on these deceptive games. Our results show that all tested agents are vulnerable to several kinds of deception, but that different agents have different weaknesses. This suggests that we can use deception to understand the capabilities of a game-playing algorithm, and game-playing algorithms to characterize the deception displayed by a game.
\end{abstract}
\section{Introduction}


\subsection{Motivation}

What makes a game difficult for an Artificial Intelligence (AI) agent? Or, more precisely, how can we design a game that is difficult for an agent, and what can we learn from doing so? 

Early AI and games research focused on games with known rules and full information, such as Chess \cite{Turing53} or Go. The game-theoretic approaches \cite{von1945theory} to these games, such as min-max, are constrained by high branching factors and large computational complexity. When Deep Blue surpassed the top humans in Chess \cite{campbell2002deep}, the game Go was still considered very hard, partly due to its much larger branching factor. Also, the design of Arimaa \cite{syed2003arimaa}, built to be deliberately difficult for AI agents, relies heavily on an even higher branching factor than Go. 

But increasing the game complexity is not the only way to make games more difficult. To demonstrate this we will here focus on old arcade games, such as Sokoban, Dig Dug or Space invaders, which can be implemented in VGDL. Part of the motivation for the development of VGDL and GVGAI was the desire to create a generic interface that would allow the same AIs to play a range of different games. GVGAI competitions have been held annually since 2013, resulting in an openly accessible corpus of games and AI agents that can play them (with varying proficiency). 

VGDL games have relatively similar game complexity: the branching factor is identical (there are six possible actions) and the game state space is not too different between games because of the similar-sized levels. Yet, if we look at how well different agents do on different games we can see that complexity is not the only factor for game difficulty. Certain games seem to be very easy, while others are nearly impossible to master for all existing agents. These effects are still present if the agents are given considerably more time which could compensate for complexity \cite{nelson2016investigating}. Further analyses also shows that games cannot easily be ordered by difficulty, as agents based on different types of algorithms seem to have problems with different games---there is a distinct non-transitivity in performance rankings~\cite{bontrager2016matching}.  This raises the question of what makes a game difficult for a specific agent but not for others? 

One way to explain this is to consider that there are several methods for constructing agents to play games. One can train a function approximator to map from a state observation to an action using reinforcement learning algorithms based on approximate dynamic programming (the temporal difference family of methods), policy gradients or artificial evolution; alternatively, and complementary, if you have a forward model of the game you can use tree search or evolution to search for action sequences that maximize some utility~\cite{yannakakis2018artificial}. Additionally, there are hybrid algorithms combining elements from several of these methods, such as the very successful AlphaGo\cite{Silver2016} system which combines supervised learning, approximate dynamic programming and Monte Carlo Tree Search.


A commonality between these game-playing methods is that they rely on rewards to guide their search and/or learning. Policies are learned to maximize the expected reward, and when a model is available, action sequences are selected for the same criterion. Fortunately, rewards are typically well-defined in games: gaining score is good, losing lives or getting hurt is bad. Indeed, one of the reasons for the popularity of games as AI testbeds is that many of them have well-defined rewards (they can also be simulated cheaply, safely and speedily). But it's not enough for there to be rewards; the rewards can be structured in different ways. For example, one of the key problems in reinforcement learning research, credit allocation, is how to assign reward to the correct action given that the reward frequently occurs long after the action was taken. 

Recently, much work has gone into devising reinforcement learning algorithms that can learn to play simple arcade games, and they generally have good performance on games that have short time lags between actions and rewards. For comparison, a game such as \emph{Montezuma's Revenge} on the Atari 2600, where there is a long time lag between actions and rewards, provides a very hard challenge for all known reinforcement learning algorithms.

It is not only a matter of the time elapsed between action and reward; rewards can be more or less helpful. The reward structure of a game can be such that taking the actions the lead to the highest rewards in the short-to-medium term leads to lower overall rewards, i.e. playing badly. For example, if you spend all your time collecting coins in \textit{Super Mario Bros}, you will likely run out of time. This is not too unlike the situation in real life where if you optimize your eating policy for fat and sugar you are likely to achieve suboptimal global nutritional reward.
Designing a reward structure that leads an AI away from the optimal policy can be seen as a form of deception, one that makes the game harder, regardless of the underlying game complexity. If we see the reward function as a heuristic function approximating the (inverse) distance from a globally optimal policy, a deceptive reward function is an \emph{inadmissible heuristic}.
\subsection{Biases, deception and optimization}

In order to understand why certain types or agents are weak against certain kinds of deceptions it is helpful to consider different types of deception through the lens of cognitive biases.  
Deceptive games can be seen as exploiting a specific cognitive bias\footnote{To simplify the text we talk about the game as if it has agency and intentions; in truth the intentions and agency lies with the game's designer, and all text should be understood in this regard.} of the (human or AI) player to trick them into making a suboptimal decision. Withholding or providing false information is a form of deception, and can be very effective at sabotaging a player's performance. In this paper though, we want to focus on games where the player or AI has full access to both the current game state and the rules (forward model). Is it still possible to design a game with these constraints that tricks an artificial agent? If we were facing an agent with unlimited resources, the answer would be no, as unbounded computational resources makes deception impossible: an exhaustive search that considers all possible action sequences and rates them by their fully modeled probabilistic expected outcome will find the optimal strategy. Writing down what a unbounded rational agent should do is not difficult. In reality, both humans and AI agents have bounded rationality in that they are limited in terms of computational resources, time, memory, etc.

To compensate for this, artificial intelligence techniques rely on approximations or heuristics that are easier to compute and still return a better answer than random. In a naive interpretation, this seems to violate the free lunch theorem. This is still a viable approach though if one only deals with a subset of all possible problems. These assumptions about the problems one encounters can be turned into helpful cognitive biases. In general, and in the right context, this is a viable cognitive strategy - one that has been shown to be effective for both humans and AI agents~\cite{tversky1974judgment,gigerenzer1996reasoning}. But reasoning based on these assumptions also makes one susceptible to deceptions - problems that violate this assumption and are designed in a way so that the, now mistaken, assumption leads the player to a suboptimal answer. Counter-intuitively, this means that the more sophisticated an AI agent becomes, the better it is at exploiting typical properties of the environment, the more susceptible it becomes to specific deceptions based on those cognitive biases.

This phenomenon can be related to the No Free Lunch theorem for search and optimization, which implies that, given limited time, making an agent perform better on a particular class of search problems will make it perform worse on others (because over all possible search problems, all agents will perform the same)~\cite{wolpert1997no}. Of course, some search algorithms are in practice better than others, because many ``naturally occurring'' problems tend to fall in a relatively restricted class where deception is limited. Within evolutionary computation, the phenomenon of deceptive optimization problems is well-defined and relatively well-studied, and it has been claimed that the only hard optimization problems are the deceptive ones~\cite{whitley1991fundamental,deb1993analyzing}.

For humans, the list of cognitive biases is quite extensive, and subsequently, there are many different deception strategies for tricking humans. Here we focus on agent which have their own specific sets of biases. Identifying those cognitive biases via deceptive games can help us to both categorize those agents, and help us to figure out what they are good at, and on what problem they should be used. Making the link to human biases could also help us to understand the underlying assumptions humans use, enabling us to learn from human mistakes what shortcuts humans take to be more efficient than AIs. 

\subsection{Overview}

The rest of this paper is structured as follows. We first outline some AI-specific deceptions based on our understanding of current game-playing algorithms. We present a non-exhaustive list of those, based on their assumptions and vulnerabilities. 
We then introduce several new VGDL games, designed to specifically deceive the existing AI algorithms. We test a range of existing agents from the GVGAI framework on our new deceptive games and discuss the results.






%
\section{Background}
%

%




\subsection{Categories of Deception}
By linking specific cognitive biases to types of deception we can categorize different deceptive games and try to predict which agents would perform well on them. We can also construct deceptive games aimed at exploiting a specific weakness. The following is a non-exhaustive list of possible AI biases and their associated traps, exemplified with some of the games we present here.
\subsubsection{Greed Trap:}
A common problem simplification is to only consider the effect of our actions for a limited future. These greedy algorithms usually aim to maximize some immediate reward and rely on the assumption that the local reward gradient will guide them to a global maximum. One way to specifically exploit this bias (a greedy trap) is to design a game with an accumulated reward and then use some initial small reward to trick the player into an action that will make a later, larger reward unattainable. The later mentioned \textit{DeceptiCoins} and \textit{Sister Saviour} are examples of this. 
Delayed rewards, such as seen in \textit{Invest} and \textit{Flower}, are 
a subtype. In that case, an action has a positive reward that is only awarded much later. This can be used to construct a greedy trap by combining it with a smaller, more immediate reward. This also challenges algorithms that want to attach specific rewards to actions, such as reinforcement learning. 
\subsubsection{Smoothness Trap:}
Several AI techniques also rely on the assumption that good solutions are ``close'' to other good solutions. Genetic Algorithms, for example, assume a certain smoothness of the fitness landscape and MCTS algorithms outperform uninformed random tree search because they bias their exploration towards branches with more promising results. This assumption can be exploited by deliberately hiding the optimal solutions close to a many really bad solutions. In the example of \textit{DeceptiZelda} the player has two paths to the goal. One is a direct, safe, low reward route to the exit which can be easily found. The other is a long route, passing by several deadly hazards but incurring a high reward if the successful route is found. Since many of the solutions along the dangerous part lead to losses, an agent operating with the smoothness bias might be disinclined to investigate this direction further, and would therefore not find the much better solution. This trap is different from the greedy trap, as it aims at agents that limit their evaluation not by a temporal horizon, but by only sampling a subset of all possible futures. 
\subsubsection{Generality Trap:}
Another way to make decision-making in games more manageable, both for humans and AI agents, is to generalize from particular situations. Rather than learning or determining how to interact with a certain object in every possible context, an AI can be more efficient by developing a generalized rule. For example, if there is a sprite that kills the avatar, avoiding that sprite as a general rule might be sensible. A generality trap can exploit this by providing a game environment in which such a rule is sensible, but for few critical exceptions. \textit{WafterThinMints} aims to realize this, as eating mints gives the AI points unless too many are eaten. So the agent has to figure out that it should eat a lot of them, but then stop, and change its behavior towards the mints. Agents that would evaluate the gain in reward greedily might not have a problem here, but agents that try to develop sophisticated behavioral rules should be weak to this deception. 

\subsection{Other deceptions}
As pointed out, this list is non-exhaustive. We deliberately excluded games with hidden or noisy information. Earlier GVGAI studies have looked at the question of robustness \cite{perez2016analyzing}, where the forward model sometimes gives false information. But this random noise is still different from a deliberate withholding of game information, or even from adding noise in a way to maximize the problems for the AI. 

We should also note that most of the deceptions implemented here are focused on exploiting the reward structure given by the game to trick AIs that are optimized for actual rewards. Consider though, that recent developments in intrinsically motivated AIs have introduced ideas such as curiosity-driven AIs to play games such as \textit{Montezuma's Revenge} \cite{bellemare2016unifying} or \textit{Super Mario} \cite{pathak2017curiosity}. The internal curiosity reward enhances the AI's gameplay, by providing a gradient in a flat extrinsic reward landscape, but in itself makes the AI susceptible to deception. One could design a game that specifically punished players for exploration.

\section{Experimental Setup} 
\label{sec:appExp}

\subsection{The GVGAI Framework}
The General Video Game AI competition is a competition focused on developing AI agents that can play real-time video games; agents are tested on unseen games, to make sure that the developer of the agent cannot tailor it to a particular game~\cite{perez20162014}. All current GVGAI games are created in VGDL, which was developed particularly to make rapid and even automated game development possible~\cite{ebner2013towards}. The competition began with a single planning track which provided agents with a forward model to simulate future states but has since expanded to include other areas, such as a learning track, a rule generation track, and a level generation track~\cite{perez2016general}.

In order to analyze the effects of game deception on GVGAI agent performance, a number of games were created (in VGDL) that implemented various types of deception in a relatively ``pure'' form. This section briefly explains the goal of each game and the reasons for its inclusion. In order to determine whether an agent had selected the rational path or not, requirements were set based on the agent's performance, which is detailed in this section also.
\subsection{DeceptiCoins (DC)}
The idea behind \textit{DeceptiCoins} is to offer agents two options for which path to take. The first path has some immediate rewards and leads to a win condition. The second path similarly leads to a win condition but has a higher cumulative reward along its path, which is not immediately visible to a short-sighted agent. Once a path is selected by the agent, a wall closes behind them and they are no longer able to choose the alternative path.

\begin{figure}[t!]
\minipage[t]{0.32\textwidth}
  \includegraphics[width=\linewidth]{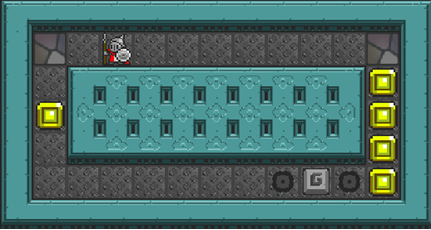}
  \caption{The first level of DeceptiCoins}\label{fig:coinLevel1}
\endminipage\hfill
\minipage[t]{0.28\textwidth}
  \includegraphics[width=\linewidth]{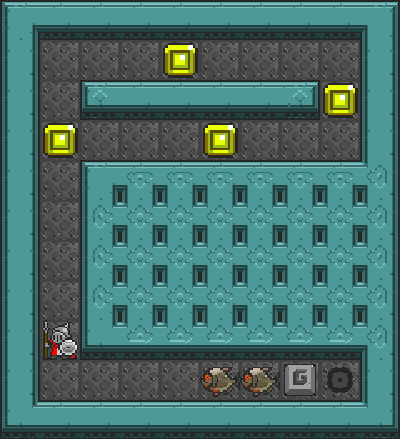}
  \caption{The second level of DeceptiCoins}\label{fig:coinLevel2}
\endminipage\hfill
\minipage[t]{0.35\textwidth}
  \includegraphics[width=\linewidth]{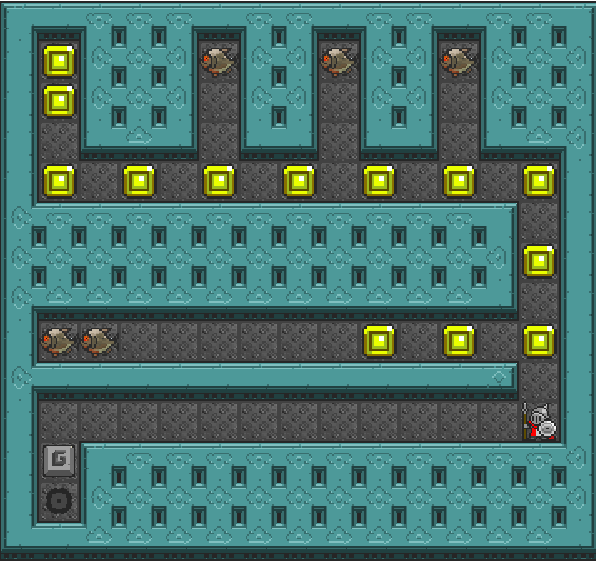}
  \caption{The third level of DeceptiCoins}\label{fig:coinLevel3}
\endminipage
\end{figure}

In order for the performance of an agent to be considered rational in this game, the agent must choose the path with the greatest overall reward. In figure \ref{fig:coinLevel1}, this rational path is achieved by taking the path to the right of the agent, as it will lead to the highest amount of score.

Two alternative levels were created for this game. These levels are similar in how the rules of the game work, but attempt to model situations where an agent may get stuck on a suboptimal path by not planning correctly. Level 2, shown in figure \ref{fig:coinLevel2}, adds some enemies to the game which will chase the agent. The agents need to carefully plan out their moves in order to avoid being trapped and losing the game. Level 3, shown in figure \ref{fig:coinLevel3} has a simple path which leads to the win condition, and a risky path that leads to large rewards. Should the agent be too greedy and take too much reward, the enemies in the level will close off the path to the win condition and the agent will lose.

The sprites used are as follows:
\begin{itemize}
\item \img{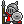} Avatar - Represents the player/agent in the game.
\item \img{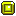} Gold Coin - Awards a point if collected.
\item \img{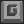} G Square - Leads to winning the game when interacted with.
\item \img{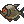} Piranha - Enemies, if the avatar interacts with these, the game is lost.
\end{itemize}

The rational paths for level 2 and 3 are defined as reaching the win condition of the level, while also collecting a minimum amount of reward (5 for level 2 and 10 for level 3).

\subsection{DeceptiZelda (DZ)}
\textit{DeceptiZelda} looks at the risk vs reward behavior of the GVGAI agents. As in \textit{DeceptiCoins}, two paths are presented to the agent, with one leading to a quick victory and the other leading to a large reward, if the hazards are overcome. The hazards in this game are represented as moving enemies which must either be defeated or avoided.

\begin{figure}[t!]
\minipage[t]{0.4\textwidth}
  \includegraphics[width=\linewidth]{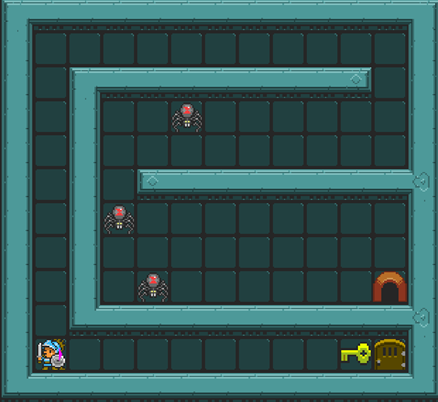}
  \caption{The first level of Deceptizelda}\label{fig:zeldaLevel1}
\endminipage\hfill
\minipage[t]{0.4\textwidth}
  \includegraphics[width=\linewidth]{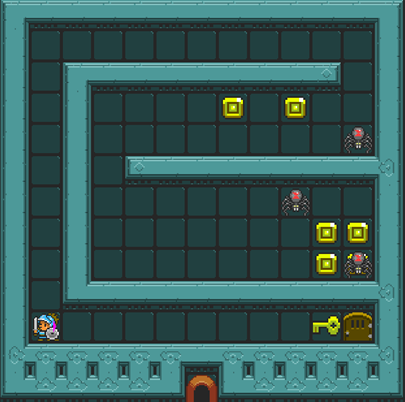}
  \caption{The second level of Deceptizelda}\label{fig:zeldaLevel2}
\endminipage
\end{figure}

Two levels for this game were created as shown in figure \ref{fig:zeldaLevel2} and figure \ref{fig:zeldaLevel1}. The first level presents the agent with a choice of going to the right, collecting the key and exiting the level immediately without tackling any of the enemies. The second path leading up takes the agent through a hazardous corridor where they must pass the enemies to reach the alternative goal. The second level uses the same layout but instead of offering a win condition, a lot of collectible rewards are offered to the agent, who must collect these and then return to the exit.

The sprites used are as follows:
\begin{itemize}
\item \img{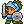} Avatar: Represents the player/agent in the game.
\item \img{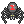} Spider: The enemies to overcome. If defeated awards 2 points.
\item \img{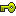} Key: Used to unlock the first exit. Awards a point if collected.
\item \img{sprites/gold1.png} Gold Coin: Awards a point to the agent if collected.
\item \img{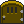} Closed Door: The low value exit. Awards a point if moved into.
\item \img{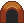} Open Door: The high value exit. Awards 10 points if moved into.
\end{itemize}

The rational path for this game is defined as successfully completing the path with the most risk. In the first level, this is defined as achieving at least 10 points and winning the game. This can be done by taking the path leading up and reaching the exit beyond the enemies. The second level of \textit{DeceptiZelda} is played on the same map, but instead of offering a higher reward win condition, a large amount of reward is available, and the agent has to then backtrack to the single exit in the level. This level can be seen in figure \ref{fig:zeldaLevel2}.

\subsection{Butterflies (BF)}

\begin{figure}[t!]
\begin{center}
\minipage[t]{0.7\textwidth}
  \includegraphics[width=\linewidth]{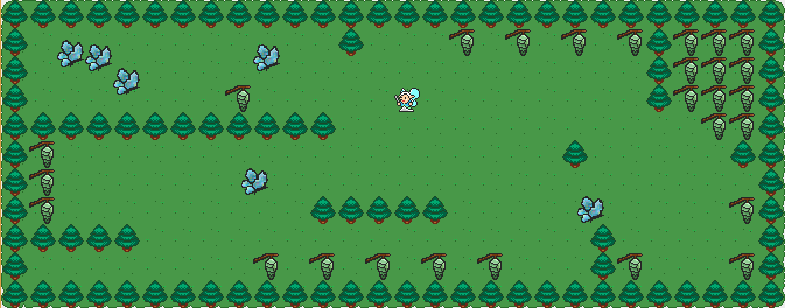}
  \caption{The first level of Butterflies}\label{fig:butterfliesLevel1}
\endminipage\hfill
\end{center}
\end{figure}

\textit{Butterflies} is one of the original games for the GVGAI that prompted the beginning of this work. This game presents a situation where if the agent aims for the win condition too quickly, they will lower their maximum potential score for the level. The goal of the game is simple; collect all of the butterflies before they reach their cocoons, which in turn creates more butterflies. To solve the game all that is required is that every butterfly is collected. Each collected butterfly grants a small reward to the agent. If the agent is able to defend a single cocoon and wait until all other cocoons have been spawned, there will be the maximum number of butterflies available to gain reward from. So long as the last cocoon is not touched by a butterfly, the game can still be won, but now a significantly higher score is possible. The level used is shown in figure \ref{fig:butterfliesLevel1}. 

The sprites used are as follows:
\begin{itemize}
\item \img{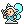} Avatar: Represents the player/agent in the game.
\item \img{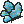} Butterfly: Awards 2 points if collected.
\item \img{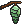} Cocoon: If a butterfly interacts with these, more butterflies are created.
\end{itemize}

The rational path for \textit{Butterflies} is defined as any win condition with a final score over 30. This is achieved by allowing more than half of the cocoons to be spawned and then winning the level.

\subsection{SisterSaviour (SS)}

The concept of \textit{SisterSaviour} was to present a moral choice to the agent. There are 3 hostages to rescue in each level, and a number of enemies guarding them, as shown in figure \ref{fig:sisterSaviourLevel1}. It is not possible for the agent to defeat these enemies immediately. The agent is given a choice of either rescuing the hostages or killing them. If the agent chooses to rescue the hostages they receive a small reward and will be able to defeat the enemies, which grants a large point reward. On the other hand, if the agent chooses to kill the hostages, they are granted a larger reward immediately, but now lack the power to defeat the enemies and will lose the game.

The sprites used are as follows:
\begin{itemize}
\item \img{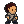} Avatar: Represents the player/agent in the game.
\item \img{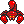} Scorpion: An enemy which chases the avatar. Immune to attacks from the avatar, unless all of the hostages have been rescued. Awards 14 points if defeated.
\item \img{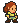} Hostage: Can be either killed, by attacking them or rescued by moving into their space. Awards 2 points if killed, and 1 point if rescued. If all are rescued then the avatar can kill the enemy.
\end{itemize}

The rational path for \textit{SisterSaviour} is defined as reaching a score of 20. This involves rescuing all of the hostages, by moving the avatar onto their space, and then defeating the enemy.

\subsection{Invest (Inv)}

\begin{figure}[t!]
\minipage[t]{0.35\textwidth}
  \includegraphics[width=\linewidth]{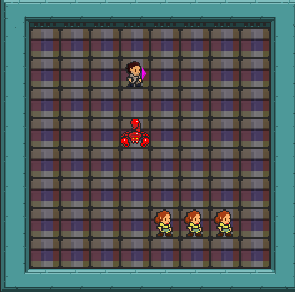}
  \caption{The first level of SisterSaviour}\label{fig:sisterSaviourLevel1}
\endminipage\hfill
\minipage[t]{0.45\textwidth}
  \includegraphics[width=\linewidth]{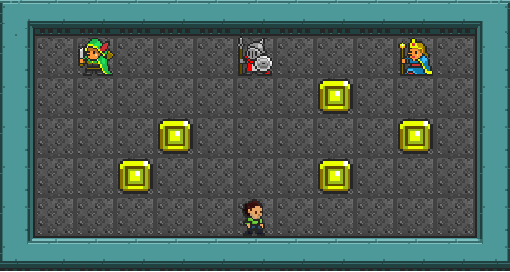}
  \caption{The first level of Invest}\label{fig:investLevel1}
\endminipage
\end{figure}

\textit{Invest} looks at the ability of a GVGAI agent to spend their accumulated reward, with the possibility of receiving a larger reward in the future. This game is shown in figure \ref{fig:investLevel1}. The agent begins with a set number of points which need to be collected from the level, which can then be spent on investment options. This is done by moving onto one of the 3 human characters to the north of the level. Investing will deduct an amount from their current score, acting as an immediate penalty, and will trigger an event to occur at a random point in the future where the agent will receive a large score reward. Should the agent invest too much, and go into a negative score, then the game is lost, otherwise, they will eventually win. The interesting point of this game was how much reward they accumulate over the time period that they have, and would they overcome any loss adversity in order to gain higher overall rewards?

The sprites used are as follows:
\begin{itemize}
\item \img{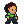} Avatar: Represents the player/agent in the game.
\item \img{sprites/gold1.png} Gold Coin: Awards a point when collected.
\item \img{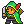} Green Investment: Takes 3 points when moved onto, returns 8.
\item \img{sprites/knight1.png} Red Investment: Takes 7 points when moved onto, returns 15.
\item \img{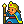} Blue Investment: Takes 5 points when moved onto, returns 10.
\end{itemize}

The rational path in \textit{Invest} is defined as investing any amount of score successfully without suffering a loss.
\subsection{Flower (Flow)}
\textit{Flower} is a game which was designed to offer small immediate rewards, and progressively larger rewards if some time is allowed to pass for the reward to grow. As shown in figure \ref{fig:flowerLevel1}, a single seed is available for the agent to collect, which is worth 0 points. As time passes the value of the seed increases as it grows into a full flower, from 0 up to 10. Once collected, the seed will begin to regrow, starting from 0 again. The rational solution for this game is to wait for a seed to grow into a full flower, worth 10 points, and then collecting it.

The sprites used are as follows:
\begin{itemize}
\item \img{sprites/man1.png} Avatar: Represents the player/agent in the game.
\item \img{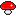} Seed: Awards 0 points initially, but this increases up to 10.
\end{itemize}

The rational path in \textit{Flower} is defined as achieving a score of at least 30. This can only be done by allowing the flower to grow to at least the second stage and consistently collecting at that level.
\subsection{WaferThinMints (Mints)}

\textit{WaferThinMints} introduces the idea that gathering too much reward can lead to a loss condition. The agent has to gather resources in order to increase their reward, but if they collect too many they will die and lose the game. 

\begin{figure}[t!]
\minipage[t]{0.32\textwidth}
  \includegraphics[width=\linewidth]{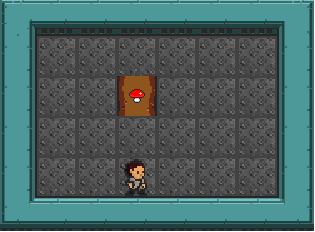}
  \caption{The first level of Flower}\label{fig:flowerLevel1}
\endminipage\hfill
\minipage[t]{0.32\textwidth}
  \includegraphics[width=\linewidth]{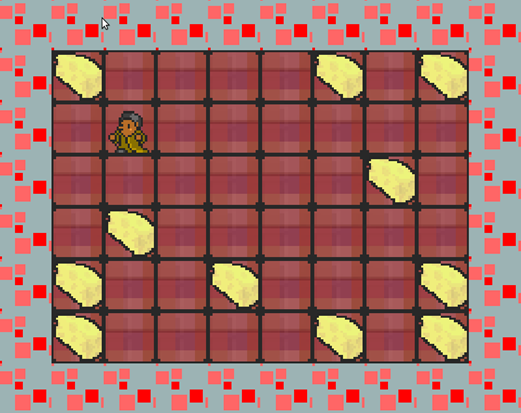}
  \caption{The first level of WaferThinMints}\label{fig:wfmLevel1}
\endminipage\hfill
\minipage[t]{0.32\textwidth}
  \includegraphics[width=\linewidth]{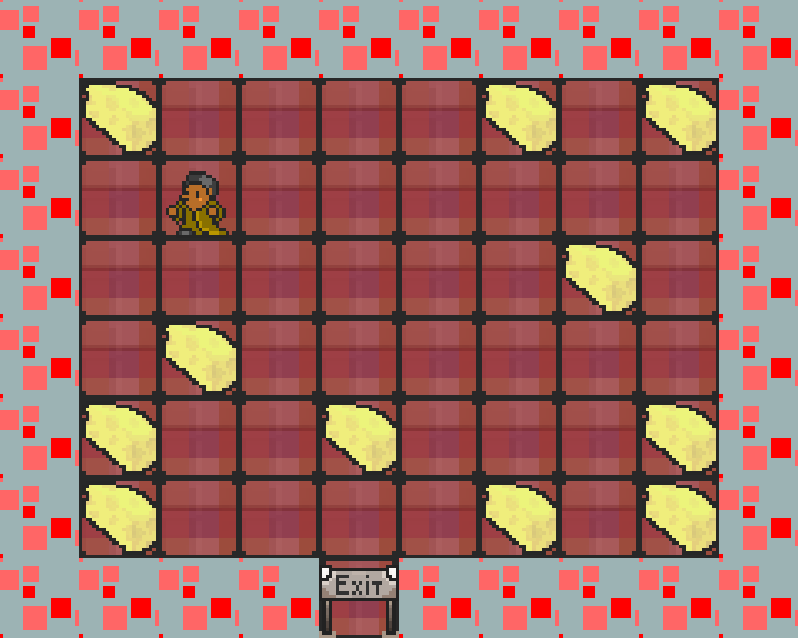}
  \caption{The second level of WaferThinMints}\label{fig:wfmLevel2}
\endminipage\hfill
\end{figure}

Two variants of this game were created. One which includes an exit from the level, shown in figure \ref{fig:wfmLevel2}, and one that does not, shown in figure \ref{fig:wfmLevel1}. These variants were created in order to provide a comparison of the effect that the deception in the level has on overall agent performance.

The sprites used are as follows: 
\begin{itemize}
\item \img{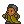} Avatar: Represents the player/agent in the game.
\item \img{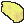} Cheese: Awards a point when collected. If 9 have been collected already, then the 10th will kill the avatar causing a loss.
\item \img{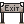} Exit: Leads to a win condition when moved into.
\end{itemize}

The rational path for both versions of the game is defined as collecting a score of 9, and then either waiting for the timeout, in level 1 or exiting the game, in level 2.

\section{Experiments and Results}
The agents used were collected from the GVGAI competitions. Criteria for selection were the uniqueness of the algorithm used and competition ranking in the past. The hardware used for all of the experiments was a Ubuntu 14.04 desktop PC with an i7-4790 CPU and 16GB Ram.

\begin{figure}[t!]
\minipage[t]{\textwidth}
  \includegraphics[width=\linewidth]{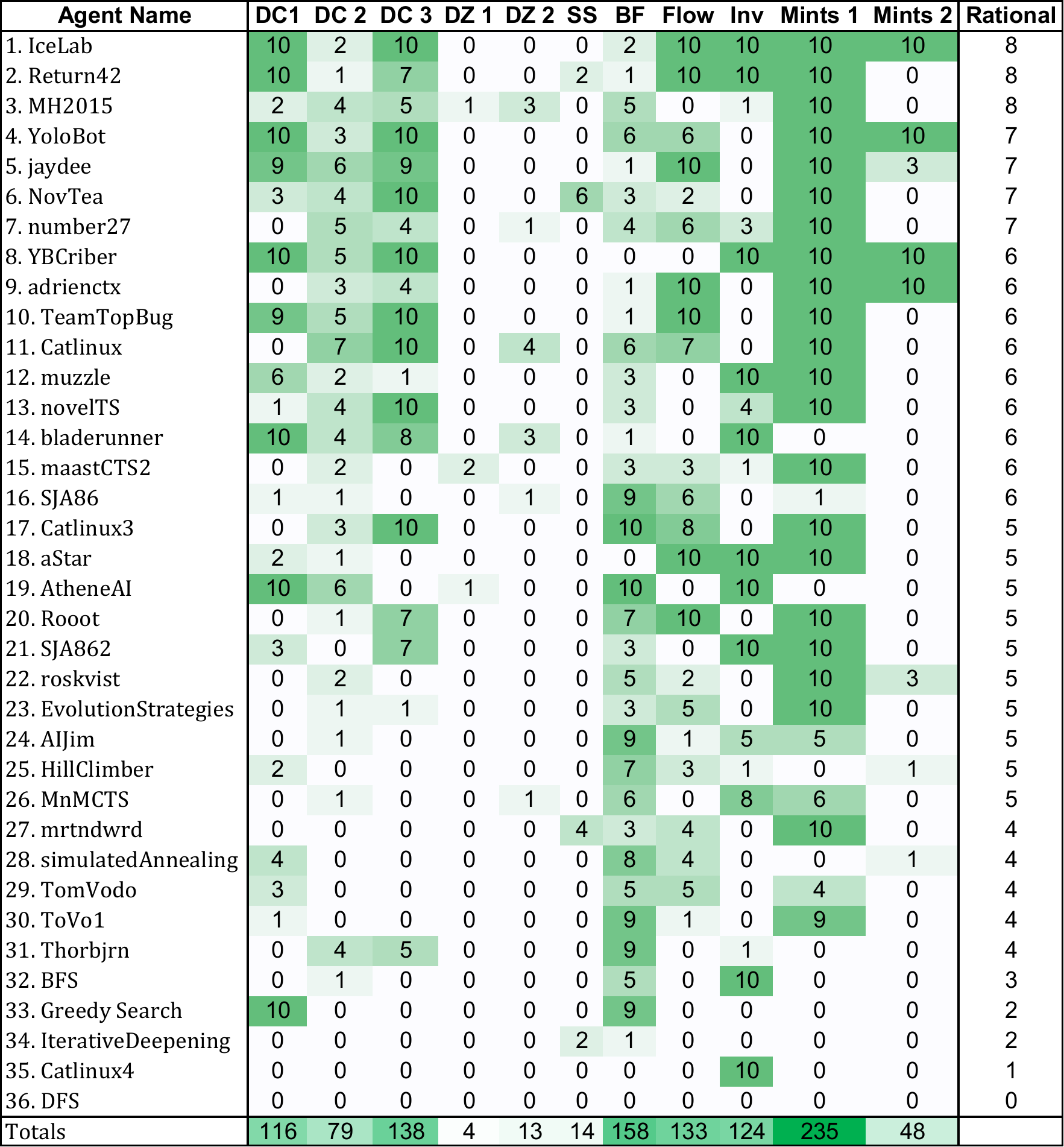}
  \caption{The results of the first experiment}\label{fig:initialResults}
\endminipage\hfill
\end{figure}

Each agent was run 10 times on each level of the deceptive games outlined in section \ref{sec:appExp}. If an agent was disqualified for any reason it was given another run to collect 10 successful results for each game and agent. In addition to comparing these performance statistics, observations were made on the choices that the agents made when faced with potentially deceptive choices. Each game's rational path is defined in section \ref{sec:appExp}. The results of these experiments are shown in table \ref{fig:initialResults}.  Each game was played a total of 360 times. The totals at the bottom of the table show how many of those games were completed using the defined rational path. The results are ranked in descending order by their number of rational trials, and then the number of games where they managed to play with 100\% rationality.

Noticeably from the initial results is that no single algorithm was able to solve all the games, with \textit{DeceptiZelda} and \textit{SisterSaviour} being particularly challenging. Furthermore, no single algorithm dominated all others in all games. For Example, IceLab, the top agent in overall results, only has 2 rational trials in \textit{Butterflies}, compared to 9 for Greedy Search, which is in the 33rd place. In general, the results for \textit{Butterflies} are interesting, as top agents perform poorly compared to some of the lower ranking agents. 

\textit{Butterflies} also has a good spread of results, with all but 4 of the algorithms being able to find the rational path at least once. While many of the algorithms are able to make some progress with the game, only 2 are able to achieve 100\% rationality.

There is an interesting difference in the performance of agents between \textit{DeceptiCoins} level 1 and 2. The agents that performed well in \textit{Decepticoins} 1 seemed to perform significantly worse in level 2. The requirements of the levels are quite different which appears to have a significant effect on the agents. If a ranking was done with only the performance of \textit{DeceptiCoins} level 2 then IceLab, the 1st ranked in this experiment, would be in the bottom half of the results table.

The hardest games for the agents to solve were \textit{DeceptiZelda} levels 1 and 2, and \textit{SisterSaviour}. \textit{DeceptiZelda’s} levels had only 4 and 13 runs solved respectively, and \textit{SisterSaviour} having 14. These games present interesting challenges to the agents, with the rational solution requiring a combination of long-range planning and sacrificing apparent reward for the superior, long-range goal.

Another interesting case here is Mints, the only game in our set with a generalization trap. Most algorithms do well in Mints, suggesting that they do not generalize. This is to be expected, as a tree search algorithm does not in itself generalize from one state to another. But bladerunner, AtheneAI, and SJA86 completely fail at these games, even though they perform reasonably well otherwise. This suggests that they perform some kind of surrogate modeling of game states, relying on a generality assumption that this game breaks. The inclusion of an accessible win condition in \textit{Mints 2} also dramatically reduced the number of algorithms that achieved the maximum amount of score, from 26 to 8. This seems to be due to also introducing a specific greed trap that most algorithms seem to be susceptible too - namely preferring to win the game outright, over accumulating more score.

Note that, the final rankings of this experiment differ quite significantly from the official rankings on the GVGAI competition. It is important to note that a different ranking algorithm is used in the competition, which may account for some of the differences observed. Many of the agents have a vastly different level of performance in these results compared to the official rankings. First of all, IceLab and MH2015 have historically appeared low in the official rankings, with their highest ranks being 10th place. The typical high ranking algorithms in the official competition seem to have been hit a bit harder by the new set of games. Yolobot, Return42, maastCTS2, YBCriber, adrienctx and number27 tend to feature in the top 5 positions of the official rankings, and have now finished in positions 2, 4, 15, 8, 9, and 7. For them to lose their positions in this new set of games could show how the games can be constructed to alter the performance of agents \cite{perez20162014,perez2016general}.

\begin{figure}[t!]
\minipage[t]{\textwidth}
\includegraphics[width=\linewidth]{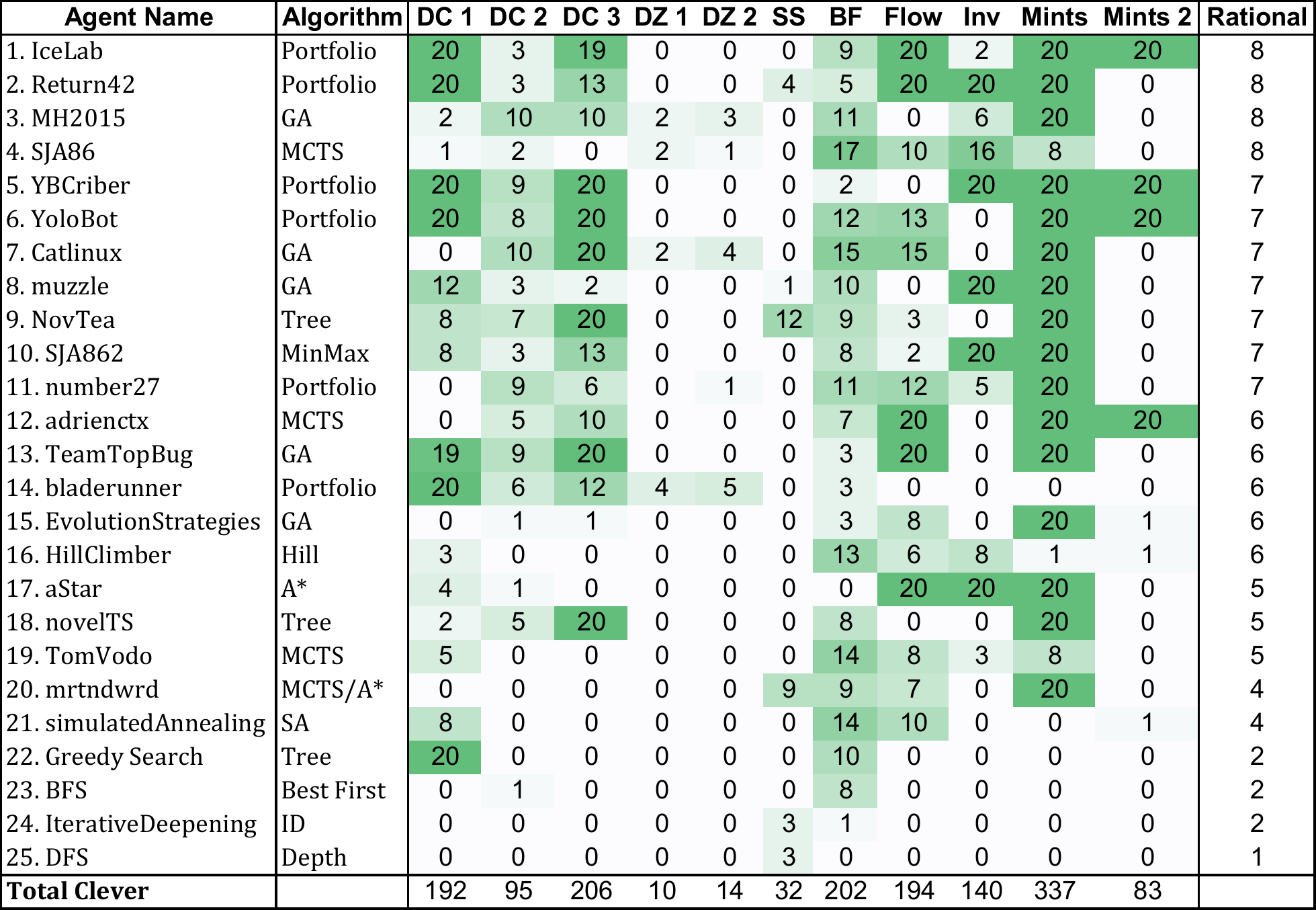}
\caption{The results of the second experiment.}\label{fig:secondResults}
\endminipage\hfill
\end{figure}

In order to look at the effect of deception on specific types of algorithms, such as genetic algorithms (GA) or Tree Search techniques, a second set of experiments were performed. A selection of algorithms were ran an additional 10 times on each of the games, and each algorithm was investigated to identify the core component of its operation. It should be noted that these classifications are simple, and an in-depth analysis of the specifics used by the algorithms might reveal some further insights. The results for these experiments are shown in figure \ref{fig:secondResults}.

These results show a number of interesting observations. First of all, for \textit{DeceptiZelda}1 and 2 it appears that agents using a genetic algorithm perform better than most other approaches, but do poorly compared to tree search techniques at \textit{SisterSaviour}. Portfolio search agents, which employ different algorithms for different games or situations, take the top two positions of the table and place quite highly overall compared to single algorithm solutions.

\section{Discussion and Future Work}

The results suggest that the types of deception presented in the games have differing effects on the performance of different algorithms. The fact that algorithms, that are more sophisticated and usually perform well in the regular competition are not on top of the rankings is also in line with our argument, that they employ sophisticated assumptions and heuristics, and are subsequently susceptible to deception. Based on the data we have now it would be possible to build a game to defeat any of the agents on the list, and it seems possible to design a specific set of games that would put any specific AI at the bottom of the table. The difficulty of a game is, therefore, a property that is, at least in part, only well defined in regards to a specific AI.   

In regards to categorization, it seems there is a certain degree of similarity between groups of games and groups of AIs that perform similarly, but a more in-depth analysis would be needed to determine what exact weakness each AI has. The games in this corpus already contain, like \textit{Mints 2}, a mixture of different deceptions. Similarly, the more sophisticated agents also employ hybrid strategies and some, like YoloBot, switch between different AI approaches based on the kind of game they detect \cite{mendes2016hyper}. One way to explore this further would be to use a genetic algorithm to create new VGDL games, with a fitness function rewarding a set of games that can maximally discriminate between the existing algorithms.

There are also further possibilities for deception that we did not explore here. Limiting access to the game state, or even requiring agents to actually learn how the game mechanics work open up a whole new range of deception possibilities. This would also allow us to extend this approach to other games, which might not provide the agent with a forward model, or might require the agent to deal with incomplete or noisy sensor information about the world. 

Another way to deepen this approach would be to extend the metaphor about human cognitive biases. Humans have a long list of cognitive biases - most of them connected to some reasonable assumption about the world, or more specifically, typical games. By analyzing what biases humans have in this kind of games we could try to develop agents that use similar simplification assumptions to humans and thereby make better agents.

\section{Acknowledgements}
Damien Anderson is funded by the Carnegie Trust for the Universities of Scotland as a PhD Scholar. Christoph Salge is funded by the EU Horizon 2020 programme under the Marie Sklodowska-Curie grant 705643.

%
%
%
%
%
%
%
\bibliographystyle{splncs03}
\bibliography{Anderson}

\end{document}